\documentclass[lettersize,journal]{IEEEtran}
\usepackage{amsmath}
\usepackage{amssymb}
\usepackage{bm}
\usepackage{mathtools}
\usepackage{verbatim}
\usepackage{tablefootnote}

\usepackage{algorithm}
\usepackage{algorithmic}

\usepackage{bbold}
\usepackage{bbm}
\usepackage{booktabs} 
\usepackage{array}  %
\usepackage[tight,normalsize,sf,SF]{subfigure}
\usepackage{multirow}
\usepackage{url}
\usepackage{color}
\usepackage{float}
\usepackage{gensymb}
\usepackage{cite}

\usepackage{tabu}
\usepackage{dcolumn}

\usepackage{amsmath,amsfonts}
\usepackage{algorithmic}
\usepackage{algorithm}
\usepackage{array}
\usepackage[caption=false,font=normalsize,labelfont=sf,textfont=sf]{subfig}
\usepackage{textcomp}
\usepackage{stfloats}
\usepackage{url}
\usepackage{verbatim}
\usepackage{graphicx}
\usepackage{cite}
\usepackage{tabu}
\usepackage{dcolumn}
\usepackage{bbold}
\usepackage{bbm}
\usepackage{booktabs} 
\usepackage{array}  %
\usepackage[tight,normalsize,sf,SF]{subfigure}
\usepackage{multirow}
\usepackage{url}
\usepackage{color}
\usepackage{float}
\usepackage{gensymb}
\usepackage{cite}

\usepackage{colortbl}
\def\mycolor{\cellcolor[rgb]{0.8275,0.8275,0.8275}}

\begin{document}

\title{LATFormer: Locality-Aware Point-View Fusion Transformer for 3D Shape Recognition}

\author{Xinwei He$^{\ast}$, 
        Silin Cheng$^{\ast}$,
        Dingkang Liang$^{\ast}$,
        Song Bai,
        Xi Wang,
        and Yingying Zhu$^{\dagger}$
        


\thanks{$^{\ast}$Authors contribute equally.}
\thanks{$^{\dagger}$Corresponding author.}
\thanks{
X. He is with the College of Informatics, Huazhong Agricultural University, Wuhan, 430070, China (e-mail: xwhe@mail.hzau.edu.cn).
}
\thanks{S. Cheng, D. Liang, and Y. Zhu are with the School of Electronic Information and Communications, Huazhong University of Science and Technology, Wuhan, 430074, China (e-mail: \{slcheng, dkliang, yyzhu\}@hust.edu.cn).}
\thanks{S. Bai is with the Department of Engineering Science, University of Oxford, Oxford, OX1 3PJ, UK (e-mail: songbai.site@gmail.com).}
\thanks{X. Wang is with CalmCar Vision Systems LLC, Beijing, 100000, China (e-mail: xi.wang@calmcar.com).}
}

\maketitle

\begin{abstract}
Recently, 3D shape understanding has achieved significant progress due to the advances of deep learning models on various data formats like images, voxels, and point clouds. Among them, point clouds and multi-view images are two complementary modalities of 3D objects and learning representations by fusing both of them has been proven to be fairly effective. While prior works typically focus on exploiting global features of the two modalities, herein we argue that more discriminative features can be derived by modeling ``where to fuse''. To investigate this, we propose a novel Locality-Aware Point-View Fusion Transformer (LATFormer) for 3D shape retrieval and classification. The core component of LATFormer is a module named Locality-Aware Fusion (LAF) which integrates the local features of correlated regions across the two modalities based on the co-occurrence scores. We further propose to filter out scores with low values to obtain salient local co-occurring regions, which reduces redundancy for the fusion process. In our LATFormer, we utilize the LAF module to fuse the multi-scale features of the two modalities both bidirectionally and hierarchically to obtain more informative features. Comprehensive experiments on four popular 3D shape benchmarks covering 3D object retrieval and classification validate its effectiveness. 
\end{abstract}


\section{Introduction}
\IEEEPARstart{3}{D} shape understanding is undoubtedly an important research topic in computer vision and has drawn increasing attention due to its essential role in a wide range of applications such as autonomous driving, virtual reality (VR), and 3D printing. 
Given a 3D object, various raw data formats (or modalities) can be employed to represent it, \emph{e.g.}, point clouds, multiple view images, or voxels. Most current works focus on learning to extract representations from only one modality, such as MVCNN~\cite{MVCNN} working on multi-view images, PointNet~\cite{PointNet} working on point cloud data, and VoxelNet~\cite{Voxnet} working on voxels. These unimodal-based methods, which adapt to a single modality of a 3D object, have led to impressive progress. 

Though significant progress has been made, these unimodal-based methods are restricted by 
the inherent representation capacity of the utilized modality. For instance, point clouds are better at capturing rich 3D geometry information of objects but may easily lose fine-grained local cues due to their sparse nature, while view images have much higher resolution and are better at describing the dense local patterns of 3D objects. It motivates us to investigate solutions to fuse representations from multiple modalities to exploit their different and usually complementary capacities. In this paper, we follow previous works~\cite{PVNet, PVRNet} and choose to fuse point cloud and multi-view data since these two modalities are direct outputs of 3D and 2D acquisition devices. 
\begin{figure}[!tb] 
\centering 
\includegraphics[width=0.96\linewidth]{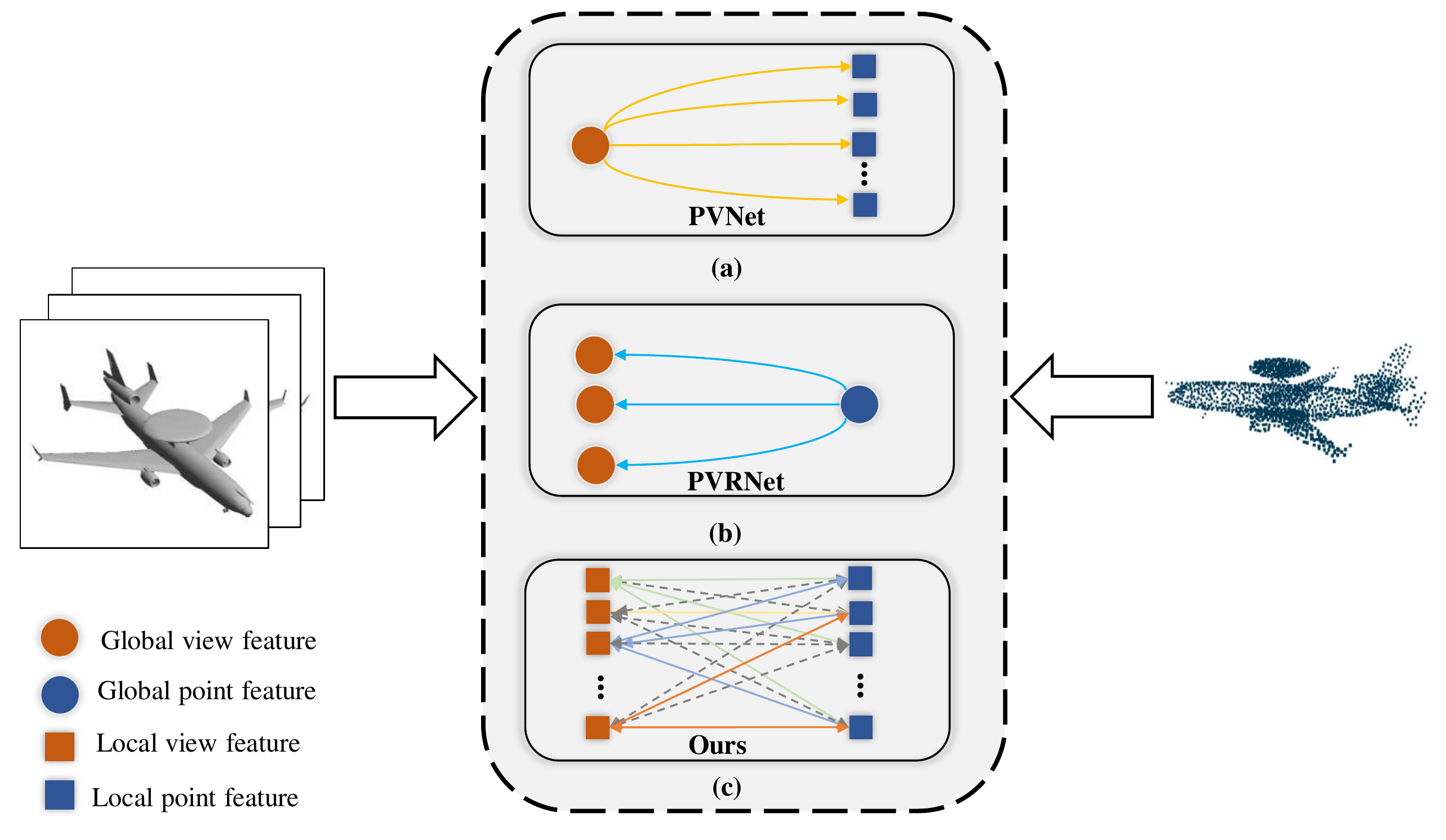} 
\caption{Overview about the fusion mechanisms of PVNet~\cite{PVNet}, PVRNet~\cite{PVRNet} and our LATFormer. Dashed lines indicate the lower relation scores across the bimodal features and are dropped during the fusion process to avoid interference of the local discriminative regions, and the arrows represent the fusion directions. 
} 
\label{Fig.diffrences_with_other_methdos} 
\end{figure}

Existing works~\cite{PVNet,PVRNet} on the fusion of point cloud and multi-view images have demonstrated that by considering the relative correlations of the bi-modal data, more discriminative 3D representations for recognition can be derived. The pioneering work PVNet~\cite{PVNet} exploits the relative correlations between the \emph{global} multi-view features and the local point cloud features, as illustrated in Fig.~\ref{Fig.diffrences_with_other_methdos} (a). PVRNet~\cite{PVRNet} further explores the benefits of learning the relations between \emph{global} features of each view and point cloud, as shown in Fig.~\ref{Fig.diffrences_with_other_methdos} (b). These earlier works have obtained promising results by fusing global features of one or both modalities. However, to differentiate 3D objects of different classes, especially the hard category pairs, the local visual details and geometric cues provide strong guidance to discriminate them. For instance, the samples from the bookshelf class and wardrobe class may have similar global visual appearances, but the local details are quite different, and directly fusing global features may dilute the discriminative local information, as is the case with PVNet and PVRNet. 

In this paper, we notice that the diverse but complementary local features\footnote{Here the local features of a 3D object are represented by the feature vectors at each pixel location of multi-view feature maps or the center point features of each local region in point clouds.} describing corresponding regions across the two modalities (shown in Fig.~\ref{Fig.diffrences_with_other_methdos}) are potentially helpful to improve the quality of the learned representations. 
For example, the recognition of an airplane by fusing the representations of co-occurrent parts (\emph{e.g.}, fins) in both modalities can preserve the rich local geometry information, which may help understand its inherent structures. However, given the multi-view images and point cloud pairs, which depict the same 3D objects, the local  features extracted are usually not aligned and thus directly concatenating them causes interference, which is another challenging problem and hinders the efficacy of the fusion process. 

To address these issues, we introduce a novel framework named Locality-Aware Point-View Fusion Transformer (LATFormer) to combine point cloud and multi-view data by fully exploiting the local region co-occurrence between the two modalities, which is presented in Fig.~\ref{Fig.diffrences_with_other_methdos} (c). Specifically, the network adopts a module named Locality-Aware Fusion (LAF) and applies it to fuse the multi-scale local structure information of the two modalities in a hierarchical manner. Unlike previous works~\cite{PVNet, PVRNet}, the LAF module selectively picks the local structure cues from one modality to enhance the relevant local region representations from the other modality. Inspired by self-attention~\cite{vaswani2017attention}, the cross-modal co-occurrence scores between all the possible local region pairs are first estimated. Then the scores with smaller values are filtered out based on a predefined threshold. The idea behind the filtering mechanism is that each local feature describing specific regions from one modality should be fused with the corresponding local features of 
similar regions to enhance the local discriminativeness, avoiding the interference of those non-correlated corresponding parts. Notably, the process of finding most correlated features for fusion has the effect of aligning the two modality inputs to some extent.
Finally, the filtered scores are used to aggregate the correlated local regions of one modality to enhance the corresponding features of another. 

One important characteristic of the LATFormer is that it fuses the point cloud and multi-view data in a bidirectional manner. Specifically, it adopts two LAFs at each scale and selects one modality as the query to be enhanced alternately. Then the remaining modality will be used to provide local cues for the query. Such a bidirectional way can ensure that each modality contributes to the final combined representations in a more interactive and comprehensive manner,  with only marginal overhead when compared with the unidirectional counterpart. 
Our LATFormer which is aware of ``where to fuse'' is very effective in combining representations of multi-view images and point clouds. 
As indicated in the ablation studies (Fig.~\ref{Fig.vis_results}), it succeeds in incorporating the most relevant local structure features from each view to enhance the corresponding anchor point features for each 3D object. More importantly, fusion based on the local co-occurrent features endows our framework with the desirable capability of handling the challenging situations of the arbitrary view setting, since the LAF module learns to localize the co-occurrent regions across the two modalities before fusion, circumventing misalignment issues. We further demonstrate that under the challenging arbitrary view setting, our method significantly outperforms its counterparts by a large margin, proving the robustness of our method. 

Another notable merit of the proposed LATFormer is that it is highly parameter-efficient, which has $2.8\times$ \emph{fewer} parameters than its bi-modal counterpart PVRNet~\cite{PVRNet}. However, encouragingly, when compared with it, we still improve the performance on the ModelNet40 dataset by 6.9\% and 0.8\% in mean Average Precision (mAP) and Overall Accuracy (OA), respectively. 

To summarize, the main contributions of this work are the following: 
\begin{itemize}
    \item We propose a novel module named Locality-Aware Fusion (LAF) to fuse representations of point cloud and multi-view data by leveraging their local spatial co-occurrence scores, which are thresholded to obtain stable complementary local features and reduce redundancy.  
    \item Based on the LAF module, we present a framework named Locality-Aware Point-View Fusion Transformer (LATFormer) to fuse multi-scale representations of the two modalities both bidirectionally and hierarchically, which is capable of mining rich cross-modal region-to-region relations, producing more expressive and robust features. 
    \item We conduct extensive experiments to analyze the proposed approach and achieve significant improvements over state-of-the-art methods on four popular 3D shape datasets.
\end{itemize}

The rest of the paper is organized as follows: we review some related works in Section~\ref{sec:related_work} and present the details of the proposed method in Section~\ref{sec:method}. Section~\ref{sec:experiments} provides the details of the experimental results and analysis. Finally, we draw conclusions in Section~\ref{sec:conclusion}.

\section{Related Work} \label{sec:related_work}
A large body of work has been proposed over the last three decades to address the problem of extracting 3D shape descriptors~\cite{saupe20013d,GIFT,fang20153d,bai2017ensemble,zhang2018inductive,TCL,HGNN}. In this section, we briefly review some recent data-driven based 3D representation learning methods. 
Generally speaking, these works can be coarsely divided into two categories: unimodal-based methods and multi-modal-based methods. 

\noindent\textbf{Unimodal based Methods.}~For unimodal based methods~\cite{MVCNN,liu2019point2sequence,PointNet,Voxnet,3DShapeNet}, they focus on learning discriminative representations from one single modality of 3D objects. The modality is often chosen among multi-view images, point clouds, voxels, or even meshes. Different modalities face different design problems in the frameworks adopted. The pioneering work on learning representations from multi-view images is MVCNN~\cite{MVCNN}. It first uses one CNN to extract the features for each view image and then performs view-pooling to aggregate a compact representation. Finally, it feeds the aggregated feature into another CNN for label prediction. However, the view-pooling layer in the MVCNN may induce information loss, especially spatial dependency. To remedy this, a lot of the following works~\cite{dai2018siamese,MHBN,GVCNN,leng2018learning,SeqViews2SeqLabels,xu2019enhancing,VNN, View-GCN} have been proposed. For instance, GVCNN~\cite{GVCNN} proposes a group-view shape description framework and View-GCN~\cite{View-GCN} employs a graph convolutional neural network to hierarchically
explore the relations of multiple views respectively. For learning from voxels, VoxNet~\cite{Voxnet} uses 3D convolutional neural networks to achieve this goal. However, since the voxel grid is memory-consuming, these methods are generally limited to low resolution (\emph{e.g.},~voxel grid of $30^3$). To alleviate this problem, OCNN~\cite{wang2017cnn} proposes to use the octree data structure. However, it still requires a serious overhead. PointNet~\cite{PointNet} is the first approach that successfully learns representations from point clouds. It uses fully-connected layers to embed the point coordinates and then pools the point-wise features together. However, it does not consider each point's local neighborhood. To address this issue, a series of works have been proposed~\cite{PointNet++,xie2018attentional,xu2018spidercnn,DGCNN, PCT, Pra-net}. For example, DGCNN~\cite{DGCNN} proposes a novel EdgeConv to extract local features by connecting each centroid with its k-nearest neighbors and achieves promising performance. Recently, several works~\cite{feng2019meshnet,hanocka2019meshcnn, lahav2020meshwalker, singh2021meshnet++} learn 3D representations directly from meshes and also obtain competitive results.  

\noindent\textbf{Multi-Modal based Methods.}~Multi-modal based methods exploit two or more modalities of 3D objects and learn to fuse them into more discriminative representations. The idea behind it is that different modalities may provide different aspects of 3D objects, and they are complementary and can be fused together to enhance the feature learning process. FusionNet~\cite{FusionNet} learns to combine representations from the volumetric and multi-view data, which brings significant
improvements to the 3D object recognition task. PVNet~\cite{PVNet} focuses on integrating global features from multi-view images with the local features from the point cloud. The following work PVRNet~\cite{PVRNet} introduces a relation score module to exploit the relationship between the point cloud and each view image. MMJN~\cite{MMJN} improves the performance by fusing panoramic view, the multi-view images, and the point cloud together. Shape Unicode~\cite{muralikrishnan2019shape} introduces a unified code to aggregate the shape cues across different modalities including voxel, point cloud, and multi-view images. CMCL~\cite{CMCL} proposes a cross-modal center loss to embed the representations of different modalities (~\emph{i.e.}, mesh, point cloud, multi-view images) into a common feature space, which aims to obtain more discriminative features for the cross-modal retrieval task.

\noindent\textbf{Transformer.}~Recently, Transformer has achieved great success in natural language processing tasks~\cite{vaswani2017attention,BERT}. This
has inspired the development of Transformer architectures for both 2D~\cite{Vit, DETR} and 3D vision applications~\cite{PCT,3DETR, hui2021pyramid,wen2022pmp}.
For instance, PointTransformer~\cite{PCT} constructs self-attention networks for various point cloud learning tasks and obtains promising results. 3DETR~\cite{3DETR} casts the 3D object detection task as a set-to-set problem and introduces a Transformer-based encoder-decoder structure to solve it with fewer hand-coded design decisions. 
In this paper, we focus on the task of 3D shape representation learning and introduce a simple but effective multi-modal fusion framework based on the transformer, which learns to combine point cloud and multi-view images both bidirectionally and hierarchically with a simple filtering strategy over local spatial co-occurrence scores. 

\section{Proposed Method}\label{sec:method} 

\begin{figure*}[!tb] 
\centering 
\includegraphics[width=0.95\textwidth,height=0.42\textwidth]{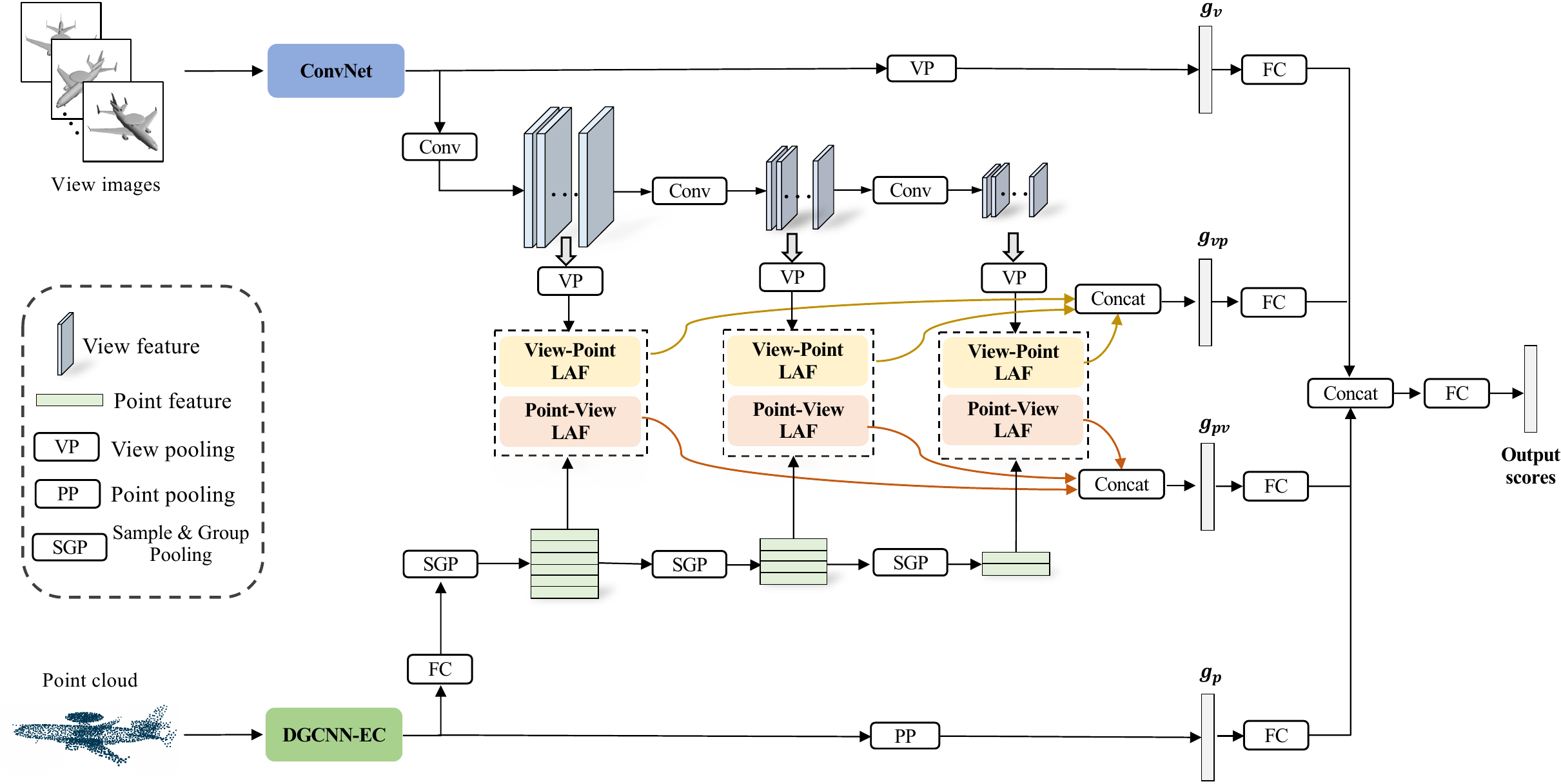} 
\caption{\textbf{Overview of LATFormer.} The Point-View LAF and View-Point LAF are both based on the LAF module (see subsection~\ref{sec.lft}). While Point-View LAF module enhances the point features by the multi-view features, the View-Point LAF enhances the multi-view features by the point features.} 
\label{Fig.architecture} 
\end{figure*}

Let $\mathcal{D}=\{(\mathcal{O}_i, y_i)\}_{i=1}^{N_o}$ denotes the training set consisting of $N_o$ 3D objects, where $\mathcal{O}_i$ represents the $i$-th 3D object and $y_i$ represents the associated category taking a value from a predefined label set. 
Without loss of generality, we discard the subscript $i$ for simplicity. 
For each 3D object $\mathcal{O}$, $N_v$ view images $\{v^{j}\}_{j=1}^{N_v}$ are projected and $N_p$ points $\{p_m\}_{m=1}^{N_p}$ are uniformly sampled from the surface to describe it, where $v^j$ represents the $j$-th projected view and $p_m \in \mathbb{R}^3$ represents the $m$-th point with the 3D coordinates as its raw feature. 
The goal of this work is to fuse representations of the two modalities (\emph{i.e.},~multi-view and point cloud) into a compact and discriminative one for 3D object retrieval and classification. To achieve this goal, we propose a simple and efficient framework named LATFormer to fuse the bi-modal inputs by leveraging their local region relations both hierarchically and bidirectionally. Fig.~\ref{Fig.architecture} illustrates the overview of the proposed LATFormer. In essence, it can be decomposed into four stages. The first two stages involve extracting multi-scale features for the multi-view images and point clouds, respectively. The third stage, which is the core part of our LATFormer, aims at fusing the bi-modal representations at each scale based on the proposed LAF module. The details of the module are elaborated in subsection~\ref{sec.lft}. Finally, the fused representations of each scale, the global point cloud and multi-view features
are combined together and fed into fully-connected (FC) layers for label prediction.
\begin{figure*}[!tb] 
\centering 
\includegraphics[width=0.9\textwidth]{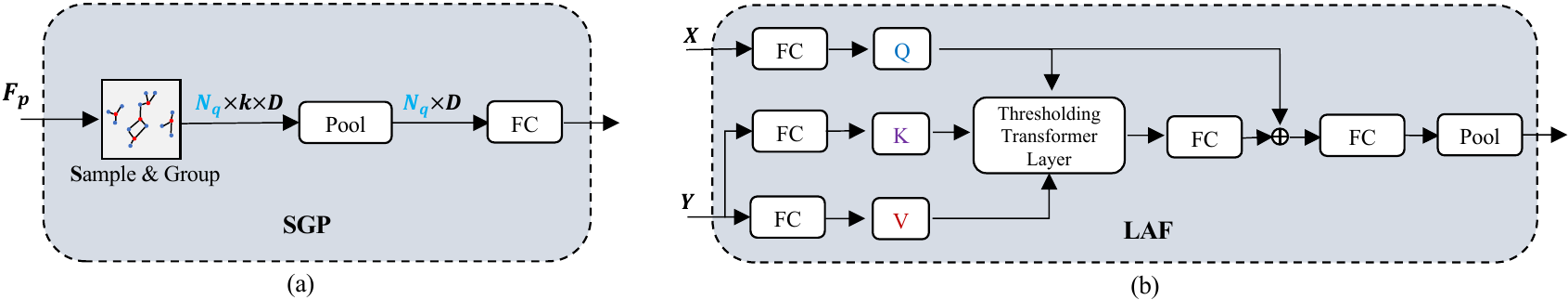} 
\caption{Structures of (a) Sample and Group Pooling (SGP) module and (b) Locality-Aware Fusion (LAF) module.} 
\label{Fig.gp_caf} 
\end{figure*}

\subsection{Extracting Multi-scale View and Point Cloud Features} 

\noindent\textbf{Extracting Multi-scale View Features.}~
Given multi-view images of a 3D object $\mathcal{O}$, we first use a stack of convolutional layers (\emph{e.g.}, all the convolutional layers in AlexNet~\cite{AlexNet} or VGG-11~\cite{VGG}) to learn region-level features. In particular, for the $j$-th view image $v^{j}$, its convolutional feature map $\mathbf{F}_v^j \in \mathbb{R}^{c \times h \times w}$ can be regarded as a sequence of
length $h \times w$, each of which is a $c$-dimensional feature representation corresponding to a specific local region in the view $v^{j}$. Based on it, we then apply $L$ convolutional layers to reduce the resolution of the feature map $\mathbf{F}_v^j$ progressively, which gives us a hierarchy of features, denoted by $\{\mathbf{F}_{v, 1}^{j}, \mathbf{F}_{v, 2}^{j}, ... , \mathbf{F}_{v, L}^{j}\}$. 
We further pack representations of the same scale from the multi-views together and perform view-pooling (\emph{i.e.},~max-pooling) to get a compact representation for each scale $l$: 
\begin{equation}
\mathbf{F}_{v, l} = \text{max-pool}(\mathbf{F}_{v,l}^1, \mathbf{F}_{v,l}^2, ..., \mathbf{F}_{v,l}^{N_v}),
\end{equation}
where $l \in \{1,2,..., L\}$. It should be noted that the feature vector at each pixel position in the feature map $\mathbf{F}_{v, l} \in \mathbb{R}^{c_v^l \times h_v^l \times w_v^l}$ gives us the most discriminative region descriptor within multi-view images. In our method, we only use three convolutional layers (\emph{i.e.},~$L=3$) for resolution reduction without changing the channel size (\emph{i.e.}, $c_v^1=c_v^2=c_v^3=D$). 
For ease of interpretation, the convolutional feature map for each scale $l$ is further flattened along the last two dimensions, producing  $\mathbf{F}_{v, l} = [\mathbf{f}_{v, l, 1}, \mathbf{f}_{v, l, 2}, ..., \mathbf{f}_{v, l, t}, ..., \mathbf{f}_{v, l, w_v^lh_v^l}]$, where $\mathbf{f}_{v, l, t} \in \mathbb{R}^{c_v^l}$.    

\noindent\textbf{Extracting Multi-scale Point Features.}~We adopt the Edge-Conv layers of DGCNN~\cite{DGCNN} (denoted as DGCNN-EConv) as the base feature extractor for the point cloud. Given a 3D point cloud $\{p_m\}_{m=1}^{N_p}$, DGCNN-EConv outputs the corresponding point-wise features $\mathbf{F}_{p} = \{\mathbf{f}_{p_m}\}_{m=1}^{N_p}$, where $\mathbf{f}_{p_m} \in \mathbb{R}^D$ represents the feature vector for point $p_m$, which is computed by aggregating its local contextual neighbors. In order to abstract the local points, inspired by PointNet++~\cite{PointNet++}, we design a module named Sample and Group Pooling (\textbf{SGP}) to perform down-sampling and feature aggregation over it. As shown in Fig.~\ref{Fig.gp_caf} (a), it first down-samples the point cloud using Furthest Point Sampling (FPS) to select $N_q$ center points. Then, for each center point $p_i$, it mines the $k$-nearest neighbors $\{p_{i,1}, p_{i,2}, ..., p_{i,k}\}$ to construct the local spatial context for it. Finally, it updates the center point features by applying max-pooling to the features of its nearest neighbor:
\begin{equation}
\mathbf{f}_{p_i}=\text{max-pool}(\mathbf{f}_{p_{i,1}}, \mathbf{f}_{p_{i,2}}, ..., \mathbf{f}_{p_{i,k}}) .
\end{equation}
In this way, we obtain a subsampled point set with $N_q$ points and their corresponding features $\mathbf{F}_{p^{\prime}} = \{\mathbf{f}_{p_i}\}_{i=1}^{N_q}$. One fully connected layer is further used to embed them. It should be noted that each center point's features can be viewed as descriptors for a local region of the original point cloud at some scale. 
In our approach, we use the \textbf{SGP} module $L$ times sequentially in order to obtain multi-scale representations of the point cloud, which give us $\mathbf{F}_{p^{\prime}, 1}$, $\mathbf{F}_{p^{\prime},2}$, ..., $\mathbf{F}_{p^{\prime}, L}$, respectively. We denote the obtained point numbers in the $L$ stages as $N_{q,1}$ ..., $N_{q,L}$. Empirically, the point numbers are decreased by a factor of 2 at each stage in order to obtain multi-scale point features. 

\subsection{Fusing Multi-view and Point Cloud Features}\label{sec.lft}

After obtaining the multi-scale representations for point cloud and multi-view images, we then propose to fuse the bi-modal representations by leveraging their rich region-to-region relations. Compared with previous methods~\cite{PVNet, PVRNet} which use the global view features or global point features for fusion, fusing the local structure features by exploiting their relevance or co-occurrence scores allows us to adaptively collect discriminative cues from each modality to learn more expressive 3D representations.
However, since there is no prior knowledge of which local regions from the two modalities are corresponding, we propose a novel module named Locality-Aware Fusion (LAF) for the fusion process. In essence, it exhaustively computes similarity scores of every local region pair across the bi-modal input and then filters out the low similarity pairs for a sparsified aggregation. The details are described in the following paragraph. 

\noindent\textbf{LAF module.}~The structure of the proposed LAF module is shown in Fig.~\ref{Fig.gp_caf} (b). As shown, the LAF module takes bi-modal inputs (denoted by $X$ and $Y$, respectively) which describe the same 3D object and learns to combine them into a compact joint representation. For ease of description, assuming $X$ and $Y$ contain $N_X$ and $N_Y$ local features, respectively, where each feature is a $D$-dimensional vector. In other words, $X \in \mathbb{R}^{N_X \times D}$ and $Y \in \mathbb{R}^{N_Y \times D}$. 
Similar to the self-attention mechanism of
transformers~\cite{vaswani2017attention}, the LAF module also includes three components: the query, the key, and the value. The query is estimated based on the modality $X$, while the key and value are computed based on the other modal input $Y$. In our design, we implement a  multi-head version for the LAF, since it allows us to further explore region relations among different representation subspaces.
Concretely, based on the bi-modal inputs $X$ and $Y$, we first apply linear layers to project them parallelly in order to obtain a sequence of $H$ tuples, which are denoted by $[(Q_1, K_1, V_1), ..., (Q_h, K_h, V_h), ..., (Q_H, K_H, V_H)]$, where $(Q_h, K_h, V_h)$ represents corresponding input tuple for $h$-th head in the form (query, key, value). The query $Q_h \in \mathbb{R}^{N_X \times D_H}$ is calculated based on $X$ while the key $K_h \in \mathbb{R}^{N_Y \times D_H}$ and value $V_h \in \mathbb{R}^{N_Y \times D_H}$ are based on $Y$.
Empirically, $D$ = $H \times D_H$. 
Then, we propose a submodule named Thresholding Transformer Layer to collect cues from $V_h$ based on the correlations between $Q_h$ and $K_h$. Mathematically, it first computes the relation or co-occurrance scores by taking the dot product of local feature at each position:
\begin{equation}
    \omega_{t, z} = Q_{h,t}K_{h,z}^T,
\end{equation}
where $\omega_{t, z}$ represents the relation score between $t$-th position in $Q_h$ and $z$-th position in $K_{h}$. The \emph{Sigmoid} function is further employed to squash the relation scores into the range (0,1). Since smaller scores indicate less related local regions across the two modalities, a threshold function is further used to filter out them and retain salient corresponding local features: 
\begin{equation}
    \alpha_{t, z} = \delta(\sigma(\omega_{t, z}), \beta), 
\end{equation}

\begin{equation}
\alpha_{t, z} =
    \begin{cases}
        \sigma(\omega_{t, z}) & \text{if } \sigma(\omega_{t, z}) > \beta\\
        0 & \text{otherwise} 
    \end{cases}
\end{equation}

where $\delta(\cdot)$ represents the threshold function, $\sigma(\cdot)$ represents the \emph{sigmoid} function taking the form $\sigma(x) = \frac{1}{1+exp(-x)}$, and $\beta$ represents the threshold value.
The analysis of the impact of the threshold $\beta$ will be presented in section~\ref{discuss}. 
Finally, we collect local features from $V_h$ based on the scores to enhance $t$-th local feature of $Q_h$ in a weighted average manner: 
\begin{equation}
    G_{h, t} = \frac{\sum_{z=1}^{N_{Y}}\alpha_{t, z} V_{h, z}}{\sum_{z=1}^{N_{Y}} \alpha_{t, z} + \epsilon}, 
\end{equation}
where $G_{h, t} \in \mathbb{R}^{1 \times D_{H}}$, and $\epsilon$ is set to 1e-5 to avoid division by zero in practice. 
After obtaining all the outputs from the multi-head branches, we concatenate them along the head dimension, \emph{i.e.},~$G = [G_1, G_2, ..., G_H]$, where $G \in \mathbb{R}^{N_X \times D}$. 
Note that the resulting features can be interpreted as the re-representation of the local features in $X$ with local features in $Y$. 
We also add a residual link to combine the bi-modal representations together by element-wise summation:
\begin{equation}
    G_{out} = \text{FC}(G) + Q, 
\end{equation}
where $\text{FC}$ is the fully connected layer, and $Q = [Q_1, Q_2, .., Q_H]$. 
Finally, The combined representation is fed into two fully connected layers, and a $\text{pooling}$ layer which can be max-pooling or mean-pooling layer\footnote{As for point features, we follow DGCNN and implement the pooling layer by concatenating max-pooling and mean-pooling layers. As for view features, we follow MVCNN and apply the max-pooling operation to the features.}  is further employed to aggregate $G_{out}$ into a compact one: 
\begin{equation}
    \mathbf{g}_{YX} = \text{pool}(\text{FC}(G_{out})),
\end{equation}
where $\mathbf{g}_{YX}$ denotes the aggregated bi-modal representation by enhancing $X$ with $Y$.
Above we have demonstrated that the representations of $X$ can be improved by collecting cues selectively from $Y$. However, it is worth mentioning that we can perform representation enhancement in both directions by simply choosing one modality for the query and the other modality for the key and value alternately. Therefore, in the same way, it is easy to obtain the enhanced features in the other direction, denoted by $\mathbf{g}_{XY}$. 
Such a bidirectional fusion strategy can integrate the local features of $X$ and $Y$ in a more interactive way. 

Specifically, in our framework, we first set $X$ and $Y$ to the local point and multi-view features, respectively, which gives us Point-View LAF. Then we switch $X$ and $Y$, which gives us View-Point LAF. %
The goal of Point-View LAF is to enhance the local point features by exploiting local multi-view features, while View-Point LAF performs feature enhancement in the other direction. 
Lastly, we insert the Point-View and View-Point LAF modules into our LATFormer to fuse the local features of the bi-modal inputs of the same scale level in a hierarchical manner.        
Recall that we have obtained a hierarchy of local features for the two modalities, respectively. For each hierarchical level $l$, we can obtain two enhanced representations with View-Point and Point-View LAF modules, which are denoted by $\mathbf{g}_{vp, l}$ and $\mathbf{g}_{pv, l}$, respectively. 
We then concatenate the outputs of different scales from $L$ View-Point LAF and Point-View LAF modules, respectively:   
\begin{equation}
\left\{ \begin{array}{ll}
        \mathbf{g}_{vp} = \mathbf{g}_{vp, 1} \oslash \mathbf{g}_{vp, 2} \oslash ... \oslash \mathbf{g}_{vp, L} \\[5mm]
        \mathbf{g}_{pv} = \mathbf{g}_{pv, 1} \oslash \mathbf{g}_{pv, 2} \oslash ... \oslash \mathbf{g}_{pv, L} 
\end{array}\right..
\end{equation}
Here, $\oslash$ denotes channel-wise concatenation. Finally, the global view features $\mathbf{g_v}$ and global point features $\mathbf{g_p}$ are combined with the above $\mathbf{g}_{vp}$ and $\mathbf{g}_{pv}$ together by concatenation after being projected through $\text{FC}$ layers respectively, which is written as:
\begin{equation}
    \mathbf{g}_{final} =\text{FC}(\mathbf{g}_{v}) \oslash \text{FC}({\mathbf{g}_{vp}}) \oslash \text{FC}(\mathbf{g}_{pv})   \oslash \text{FC}(\mathbf{g}_{p}) .
\end{equation}
Then a network with three fully connected layers is further used to make a prediction on the category for the given 3D object. We use cross-entropy loss to train our model.

\section{Experiments} \label{sec:experiments}
\subsection{Datasets and Evaluation Metrics}

We conduct experiments on four popular benchmarks for 3D shape retrieval and classification, as described below:

\noindent\textbf{ModelNet40.} The ModelNet40 dataset~\cite{ModelNet} includes 12,311 CAD models divided into 40 categories. Among them, 9,843 models are used for training, and 2,468 are for testing. 

\noindent\textbf{ModelNet10.} The ModelNet10 dataset~\cite{ModelNet} is another widely-used 3D object dataset consisting of 4,899 3D models in 10 categories. The training set and testing set have 3,991 and 908 models, respectively. 

\noindent\textbf{3D-FUTURE.} 3D-FUTURE~\cite{3D-FUTRUE} is a recent richly-annotated dataset that supports 3D shape understanding. It contains 9,992 3D furniture shapes distributed in 34 categories. It has 7 super-categories, each of which has 1-12 sub-categories at a fine-grained level. Its official train-test split includes 6,699 and 3,293 samples, respectively. This dataset is challenging due to its fine-grained furniture categories in the household scenario, requiring more detailed structure information to differentiate. 

\noindent\textcolor{black}{\textbf{ScanObjectNN.} ScanObjectNN~\cite{uy2019revisiting} includes 15, 000 real scanned objects divided into 15 classes with 2, 902 unique object instances. Compared with synthetic data, objects in this data are from real-world scenes with occlusions and background noise. 
Therefore, it poses more challenges for accurate recognition. Following previous works~\cite{qian2022pointnext, ma2021rethinking}, we conduct experiments on PB\_T50\_RS, which is the hardest and most commonly used variant.}

For the 3D shape classification experiments, we adopt Overall Accuracy (OA) and mean class Accuracy (mAcc) as the evaluation metrics. For the 3D shape retrieval task, mean Average Precision (mAP) is reported. 

\subsection{Implementation Details} 

Unless specified, we use the same preprocessing scheme as PVRNet~\cite{PVRNet} in our experiments, which uniformly samples 1,024 points and projects 12 views for each 3D model. The base feature extractors for the two modalities, \emph{i.e.}, DGCNN-EConv and ConvNet, are initialized by the pretrained DGCNN and MVCNN like PVRNet. Specifically, we adopt AlexNet and VGG-11 as the ConvNet for the ModelNet subsets under the setup of MVCNN~\cite{MVCNN} and RotationNet~\cite{kanezaki2018rotationnet} respectively; while for 3D-FUTURE, VGG-11 is employed. In our experiments, $D$ is 256. 
On the ModelNet40 (\textit{resp}., 3D-FUTRUE) dataset, the resulting three scale feature maps for the multi-view images are of size $256 \times 6 \times 6$ (\textit{resp}., $256 \times 7 \times 7$), $256 \times 4 \times 4$ (\textit{resp}., $256 \times 5 \times 5$) and $256 \times 2 \times 2$ (\textit{resp}., $256 \times 3 \times 3$), respectively.
For experiments on ScanOBjectNN~\cite{uy2019revisiting}, we project each point cloud into $10$ depth maps of size $224 \times 224$  online following previous works~\cite{zhu2022pointclip}.
We utilize AlexNet as the backbone for the depth maps and PointNeXt~\cite{qian2022pointnext} as the backbone for point clouds. 

For optimization, we first fix the parameters of ConvNet and DGCNN-EConv, and train the fusion layers with a Learning Rate (LR) of 1e-2 and a momentum of 0.9 for training stability. The LR is then decayed by a factor of 0.5 every 3 epochs.
After 6 epochs, we set the LR to 1e-3 and fine-tune the whole model. The LR is further decayed by a factor of 0.5 every 10 epochs. For the retrieval experiments, the penultimate layer output, which is 256-dimensional, is used as the descriptor for each 3D shape. 

\subsection{Shape Retrieval}

To evaluate the learned representations of 3D objects, we conduct experiments on 3D object retrieval on ModelNet40 and ModelNet10. The comparison results are summarized in Table~\ref{tab:modelnet_retrieval}. 

\begin{figure*}[!tb] 
\centering 
\includegraphics[width=\linewidth]{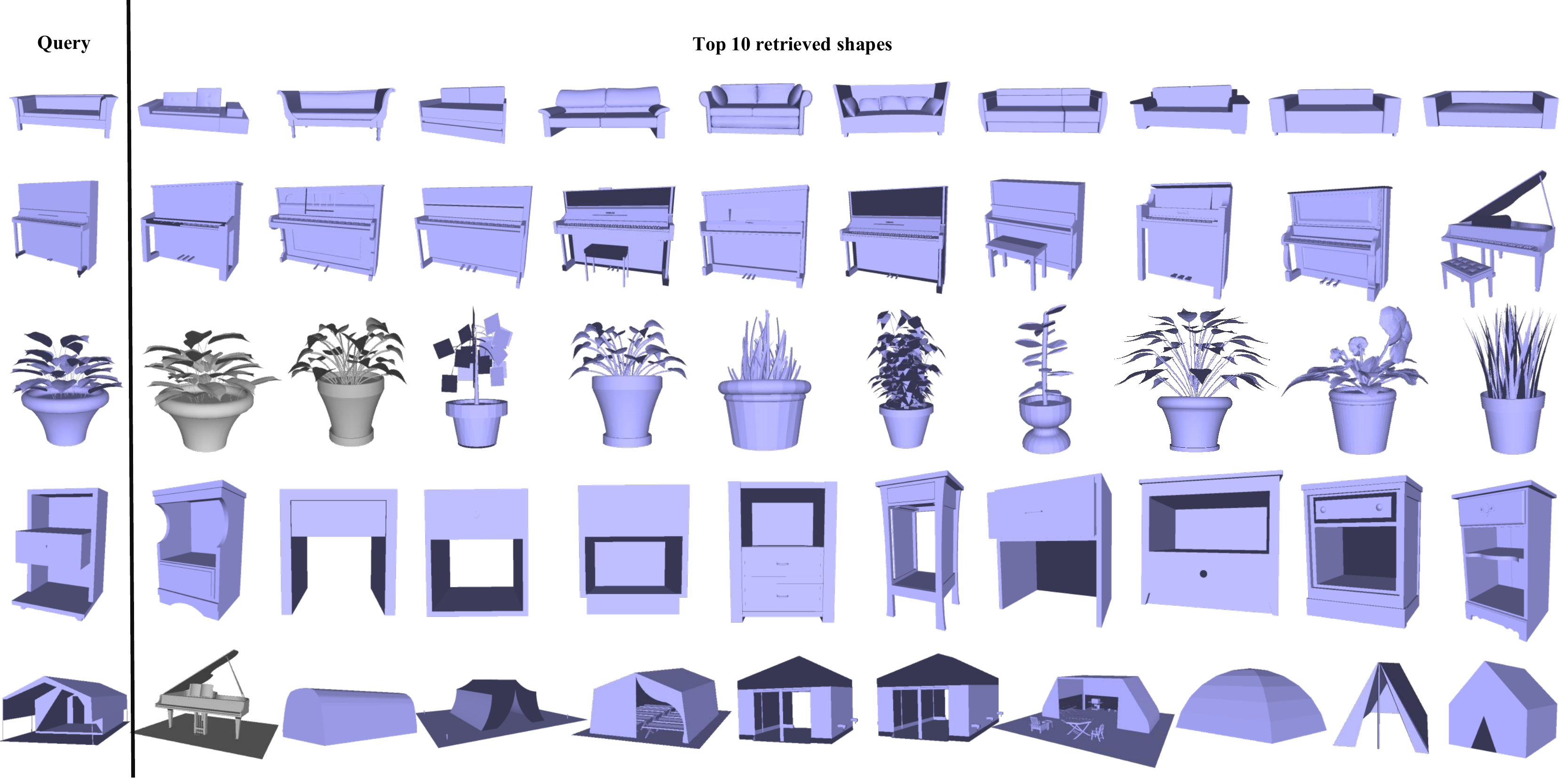} 
\vspace{1ex}
\caption{Retrieval examples of our method. Each row illustrates the top 10 retrieval shapes for the query (left-most). The grey color indicates false positives.} 
\vspace{-2ex}
\label{Fig.retrieval_vis} 
\end{figure*}

On ModelNet40, LATFormer surpasses all methods by a large margin. It is worth mentioning that the MVCNN has adopted metric learning over the learned representation to boost the retrieval performance, while we directly take the representations from our framework as the descriptor for 3D object retrieval. However, compared with MVCNN, we significantly surpass it by 12.9\% in mAP. MRVA-Net obtains the best mAP of all the unimodal-based methods, reaching 95.5\% in mAP. We improve upon it by 1.9\%. 
Among the multi-modal-based methods, CMCL achieves the best performance, reaching 91.8\% in mAP. Compared with it, we gain an improvement of 5.6\%. In Figure~\ref{Fig.retrieval_vis}, we show some retrieval examples of our method on the ModelNet40 dataset. It can be observed that our method can retrieve quite similar 3D objects for the queries. Noticeably, we believe that our approach has retrieved the correct samples for the query of the flower pot class (third row). However, due to the annotating ambiguity in the test set, the returned results are assumed to be incorrect. 

On the ModelNet10 dataset, the proposed method once again demonstrates its superiority with the best mAP of 98.7\%. DensePoint~\cite{Densepoint}, which aims at learning densely contextual representation for point clouds, has achieved the best mAP of 93.2\%. Compared with it, LATFormer outperforms it by 5.5\%. The Improved MVCNN~\cite{Improved_MVCNN} obtains the best performance among the view-based methods on ModelNet40, reaching 93.0\% in mAP, we improve upon it by 5.7\%. 
All these results convincingly demonstrate the superiority of LATFormer.

\begin{table}[t]
\small
\centering
\caption{Retrieval results in mAP (\%) on ModelNet.
The results of point-based methods are taken from ~\cite{Densepoint} and ~\cite{PointAugment}.
}
\label{tab:modelnet_retrieval}
\vspace{-2ex}
\begin{tabular}{l p{0.9cm}<{\centering} p{1.4cm}<{\centering} p{1.4cm}<{\centering}}
\toprule
Method & Input &ModelNet40&ModelNet10\\
\midrule
PointNet~\cite{PointNet} &$P_t$& 70.5 & -\\
PointNet++~\cite{PointNet++} &$P_t$& 81.3 & -\\
DGCNN~\cite{DGCNN} &$P_t$& 85.3 & -\\
PointCNN~\cite{PointCNN} &$P_t$& 83.8 & -\\
DensePoint~\cite{Densepoint} &$P_t$& 88.5 & 93.2\\
\midrule
MVCNN~\cite{MVCNN} &$I$ & 80.2 & -\\
GIFT~\cite{GIFT} &$I$& 81.9 & 91.1\\
SeqViews2Seqlabels~\cite{SeqViews2SeqLabels} &$I$& 89.1 & 91.4\\
TCL~\cite{TCL} &$I$& 88.0 & -\\
VNN~\cite{VNN} &$I$& 88.9 & 92.8\\
RN~\cite{RN} &$I$& 86.7 & -\\
3DViewGraph~\cite{3Dviewgraph} &$I$& 90.5 & 92.4\\
Improved MVCNN~\cite{Improved_MVCNN}&$I$ & 90.1 & 93.0\\
SVHAN~\cite{SVHAN} &$I$& 90.9 & 92.7\\
LIFN~\cite{zhu2022local} &$I$& 93.6 & -\\
3D2SeqViews~\cite{han20193d2seqviews} &$I$& 90.8 & 92.1\\
MRVA-Net~\cite{lin2023multi}  &$I$& 95.5 & - \\
\midrule
PVNet~\cite{PVNet} &$P_t$,$I$& 89.5 & -\\
PVRNet~\cite{PVRNet} &$P_t$,$I$& 90.5 & - \\
MMJN~\cite{MMJN} &$P_t$,$I$,$P_n$ & 89.8 & -\\
CMCL~\cite{CMCL} &$P_t$,$I$,$M_s$& 91.8 & - \\
\mycolor{LATFormer} &\mycolor{$P_t$,$I$}&\mycolor{\textbf{97.4}} & \mycolor{\textbf{98.7}} \\
\bottomrule
\end{tabular}
\end{table}

\subsection{Shape Classification}

\noindent\textbf{3D-FUTURE.} The 3D-FUTURE dataset poses great challenges to existing 3D shape classification techniques due to its hierarchical structure and fine-grained attributes of each object. The classification results are presented in Table~\ref{tab:3D-FUTURE}. As shown, our approach outperforms all the unimodal based methods by a large margin. Particularly, it significantly surpasses the best view-based method, \emph{i.e.}, GVCNN~\cite{GVCNN} with an increase of 1.8\% in OA and 3.7\% in mAcc. When compared with the best point-based method, \emph{i.e.}, DGCNN~\cite{GVCNN}, we obtain 1.4\% and 1.9\% improvements in terms of OA and mAcc, respectively. The superior performance on the challenging industrial CAD model dataset indicates that our method has great potential in practical applications.

\begin{table}[t]
\small
\centering
\setlength{\tabcolsep}{4mm}
\caption{Classification results on 3D-FUTURE dataset. $P_t$ represents point cloud, $I^{\ddagger}$ represent view images under setup of ~\cite{3D-FUTRUE}. The results from point-based methods are taken from ~\cite{Densepoint} and ~\cite{PointAugment}. $^*$ indicates results reproduced in our environment, while other results are taken from~\cite{3D-FUTRUE}.}
\label{tab:3D-FUTURE}
\begin{tabular}{ lccc }
\toprule
Method &Input & OA&mAcc\\
\midrule
MVCNN~\cite{MVCNN} &$I^{\ddagger}$& 69.2 & 65.4\\
GVCNN$^*$~\cite{GVCNN} &$I^{\ddagger}$& 70.8 & 66.7 \\
View-GCN$^*$~\cite{View-GCN}&$I^{\ddagger}$ & 70.4 & 68.0\\
\midrule
PointNet$^*$~\cite{PointNet} &$P_t$& 68.8 & 66.2 \\
PointNet++~\cite{PointNet++} &$P_t$& 69.9 & 66.0\\
DGCNN$^*$~\cite{DGCNN}&$P_t$ & 71.2 & 68.5 \\
\midrule
\mycolor{LATFormer} &\mycolor{$P_t$, $I^{\ddagger}$}&\mycolor{\textbf{72.6}} & \mycolor{\textbf{70.4}}\\
\bottomrule
\end{tabular}
\end{table}

\noindent\textbf{ScanObjectNN.}
\textcolor{black}{In Table~\ref{tab:scanobjectnn}, we compare the performance on ScanObjectNN~\cite{uy2019revisiting} with representative methods. 
Compared with the best depth-view-based method SimpleView++~\cite{sheshappanavar2022simpleview++}, we outperform it by 3.9\% in OA.  
PointNeXt~\cite{qian2022pointnext} is a recent state-of-the-art method, reaching a strong performance of 88.2\% and 86.8\% in OA and mAcc, respectively. 
PointNeXt+HyCoRe~\cite{montanaro2022rethinking} improves it with regularization in the hyperbolic space, further advancing the OA and mAcc by 0.1\% and 0.2\% in OA and mACC, respectively. Compared with it, we obtain 0.4\% and 0.1\% improvements in terms of OA and mAcc
Compared with the most recent PointVector-S~\cite{deng2023pointvector} and PointMetaBase-S~\cite{lin2023meta}, we outperform them by 0.5\% and 0.6\% in OA, respectively. 
It again proves that LATFormer is effective at fusing features from point clouds and images in real-world settings where background noise and occlusions exist.}

\begin{table}[t]
\small
\centering
\setlength{\tabcolsep}{4mm}
\caption{Classification results on ScanObjectNN. $P_t$ represents point cloud, $I_d$ represent depths images, and \_ means unknown.}
\label{tab:scanobjectnn}
\begin{tabular}{ lccc }
\toprule
Method &Input & OA&mAcc\\
\midrule
SimpleView~\cite{goyal2021revisiting} &$I_{d}$& 80.5 & - \\
SimpleView+~\cite{sheshappanavar2022simpleview++} &$I_{d}$& 84.8	& - \\
\midrule
PointNet++~\cite{PointNet++} &$P_t$& 77.9 & 75.4 \\ 
DGCNN~\cite{DGCNN}&$P_t$ & 78.1 & 73.6 \\ 	
PointMLP~\cite{ma2021rethinking} &$P_t$& 85.7 & 84.4 \\  	
PRA-Net~\cite{cheng2021net} &$P_t$ & 82.1 & 79.1\\ 	
PointNeXt~\cite{qian2022pointnext}&$P_t$ & 88.2 & 86.8 \\		
PointNeXt+HyCoRe~\cite{montanaro2022rethinking}&$P_t$ &  88.3 & 87.0 \\
PointVector-S~\cite{deng2023pointvector}&$P_t$ &  88.2 & 86.7 \\ 
PointMetaBase-S~\cite{lin2023meta}&$P_t$ &  88.1 & - \\
\midrule
\mycolor{LATFormer} &\mycolor{$P_t$, $I_{d}$}&\mycolor{\textbf{88.7}} & \mycolor{\textbf{87.1}}\\
\bottomrule
\end{tabular}
\end{table}

\noindent\textbf{ModelNet.} The comparison results with state-of-the-art methods on 3D shape classification are summarized in Table~\ref{tab:modelnet_classification}. As shown, the proposed method outperforms all previous methods with 97.9\% in OA and 97.0\% in mAcc on the ModelNet40 dataset. Compared with the unimodal based methods like MVCNN-MultiRes, CurveNet, and View-GCN, our LATFormer improves upon them by 4.1\%, 4.1\%, and 0.3\% in OA, respectively. We also compare with point-view based methods, including PVNet and PVRNet. For fair comparisons, we adopt the same rendering techniques and point-view backbones with ~\cite{PVNet, PVRNet}, and our LATFormer outperforms them by more than 0.8\% in OA, respectively. What's more, compared with MMJN, which utilizes three modalities for fusion, our method still surpasses it by 0.6\% in OA. Besides, we also evaluate our method on the ModelNet10 dataset. It again significantly outperforms all the compared methods in terms of both OA and mAcc, reaching 99.1\% and 99.0\%, respectively.
The results on both ModelNet40 and ModelNet10 consistently demonstrate the competitiveness of our method on fusing point clouds and view images of 3D objects.

\begin{table}[H]
\small
\centering
\setlength{\tabcolsep}{1.2mm}
\caption{Classification results on ModelNet40 and ModelNet10 datasets. $V_x$, $P_t$ and $P_n$ represent voxels, point clouds and panorama-views, respectively. $I^{\diamond}$ and $I^{\dagger}$ represent view images under setup of ~\cite{MVCNN} and ~\cite{kanezaki2018rotationnet} respectively.}
\label{tab:modelnet_classification}
\vspace{-2ex}
\begin{tabular}{l p{1.4cm}<{\centering} *{2}{p{0.7cm}<{\centering}} *{2}{p{0.7cm}<{\centering}}}
\midrule
\multirow{2}{*}{Methods}& \multirow{2}{*}{Inputs} & \multicolumn{2}{c}{ModelNet40} & \multicolumn{2}{c}{ModelNet10}  \\
\cline{3-4} \cline{5-6} 
    &   & OA & mAcc  & OA & mAcc  \\
\midrule
3D shapeNets~\cite{ModelNet} &$V_x$ & - & 77.3 & - & 83.5 \\
VoxNet ~\cite{Voxnet} &$V_x$ &  - & 83.0 & - & 92.0\\
MVCNN-MultiRes ~\cite{MVCNN_MultiRes} &$V_x$ &  93.8 & 91.4 & - & -\\
\midrule
PointNet~\cite{PointNet} &$P_t$ & 89.2 & 86.2 &91.9 &-\\
PointNet++~\cite{PointNet++} &$P_t$ & 91.9 & 90.2 &93.3 &-\\
Kd-Networks~\cite{Kd-Networks} &$P_t$ & 91.8 & 88.5 & 94.0 & 93.5\\
DGCNN~\cite{PointNet++} &$P_t$ & 92.9 & 90.2 &94.8 &-\\
CurveNet~\cite{xiang2021walk} &$P_t$ & 93.8 & - & 96.1 &-\\
\midrule
DeepPano~\cite{MVCNN} &$P_n$ & 77.6 & - &85.5&-\\
MVCNN~\cite{MVCNN} &$I^{\diamond}$ & 90.1 & 90.1 &-&-\\
GVCNN~\cite{GVCNN} &$I^{\diamond}$ & 93.1 & 90.7 &-&-\\
MHBN~\cite{MHBN} &$I^{\diamond}$ & 94.1 & 92.2 & 94.9 & 94.9\\
3D2seqViews~\cite{3D2SeqViews} &$I^{\diamond}$ & 93.4 & 91.5 & 94.7 & 94.7\\
SeqView2SeqLabels~\cite{SeqViews2SeqLabels} &$I^{\diamond}$ & 93.3 & 91.1 & 94.8 & 94.8\\
RotationNet~\cite{kanezaki2018rotationnet} &$I^{\dagger}$ & 97.4 & - & 98.5 & -\\
View-GCN~\cite{View-GCN} &$I^{\dagger}$ & 97.6 & 96.5 & - & -\\
\midrule
FusionNet~\cite{FusionNet} &$V_x$, $I^{\diamond}$ & 90.8 & - & 93.1 & -\\
PVNet~\cite{PVNet} &$P_t$, $I^{\diamond}$ & 93.2 & - & - & -\\
PVRNet~\cite{PVRNet} &$P_t$, $I^{\diamond}$ & 93.6 & - & 93.8 & -\\
MMJN~\cite{MMJN} &$P_t$, $I^{\diamond}$, $P_n$ & 93.8 & 92.2 & 93.8 & -\\
\mycolor{LATFormer} &\mycolor{$P_t$, $I^{\diamond}$} & \mycolor{\textbf{94.4}} & \mycolor{\textbf{92.2}} & \mycolor{\textbf{95.9}} & \mycolor{\textbf{95.8}}\\
\mycolor{LATFormer} & \mycolor{$P_t$, $I^{\dagger}$} &\mycolor{\textbf{97.9}} & \mycolor{\textbf{97.0}} & \mycolor{\textbf{99.1}} & \mycolor{\textbf{99.0}}\\
\midrule
\end{tabular}
\end{table}

\subsection{Discussions} \label{discuss} 

In this subsection, we give an in-depth analysis of the proposed LATFormer. For convenience, ModelNet40 is employed for all ablation experiments.We use the same preprocessing scheme for the view images as PVRNet~\cite{PVRNet} unless otherwise specified. 

\noindent\textbf{Robustness to Arbitrary View Setup.} In this setting, we utilize VGG-11 and DGCNN as the feature extractor of view and point cloud branch, respectively. Table~\ref{tab:Arbitary} summarizes the performance of our method under the challenging arbitrary view setting~\cite{CVR}. This setup brings great challenges like unaligned input, causing a significant performance drop of existing methods when compared to their aligned counterparts, \emph{e.g.}, the OA drops greatly by over 10\% for MVCNN and CVR. While under this case, our method suffers the least performance drop, outperforming the view-based method MVCNN and CVR by 9.3\% and 5.9\% respectively, and the point-view fusion method PVRNet by 3.8\% in OA. We consider that it is because 1) incorporating extra modality (\emph{i.e.}, point cloud) into the framework can improve the robustness to the challenging view-alignment issue, 2) aggregating based on the local co-occurrent regions across the two-modality has the effect of aligning the features across the two-modalities before fusion, therefore more robust representations are obtained.
\begin{table}[t]
\small
\centering
\caption{Classification results under arbitrary view setup~\cite{CVR}. $^*$ indicates results reproduced in our
environment, while other results are taken from ~\cite{CVR}.}
\label{tab:Arbitary}
\vspace{-2ex}
\begin{tabular}{l *{2}{p{0.7cm}<{\centering}} *{2}{p{1.7cm}<{\centering}}}
\midrule
\multirow{2}{*}{Methods}& \multicolumn{2}{c}{Aligned} & \multicolumn{2}{c}{Arbitary} \\
  \cline{2-5}
  &OA & mAcc & OA & mAcc\\
\midrule
MVCNN~\cite{MVCNN}  & 96.4 & 94.3 & 83.2 ($\downarrow$ 13.2) & 78.9 ($\downarrow$ 15.4) \\
CVR~\cite{CVR}  & 97.2 & 95.8 & 86.9 ($\downarrow$ 10.3) & 84.0 ($\downarrow$ 11.8)\\
\midrule
PVRNet$^*$~\cite{PVRNet}  & 97.6 & 96.2 & 89.0 ($\downarrow$ 8.6) & 83.9 ($\downarrow$ 12.3)\\
LATFormer  & \textbf{97.9} & \textbf{97.0} & \textbf{92.8} ($\downarrow$ 5.1) & \textbf{89.5} ($\downarrow$ 7.5)\\
\bottomrule
\end{tabular}
\end{table}

\noindent\textbf{Influence of fusion strategies.}~Table~\ref{tab:different fusion strategies} compares our method with other baseline fusion strategies. ``Late Fusion'' represents the fusion method by simply concatenating the global features from the point cloud and multi-view data. ``Our DeepConcat'' represents fusion by concatenating the hierarchy of the local features from the two modalities. ``View-Point Fusion'' represents the fusion method by only using View-Point LAF modules. ``Point-View Fusion'' represents the fusion method by only using Point-View LAF modules. As shown, simply concatenating global features of the two modalities can improve a little over unimodal based methods on the classification task. However, our DeepConcat improves upon it by 0.7\% in mAcc and 1.0\% in OA, which proves the benefits of fusion with local features. More importantly, our LATFormer increases its OA from 93.6\% to 94.4\% and the mAcc from 91.5\% to 92.2\%, which suggests the benefits of exploiting the local region relations of the two modalities in the fusion process. Besides, we also find that the bidirectional learning indeed produces better results than the unidirectional learning from the comparison between LATFormer with View-Point Fusion or Point-View Fusion strategy. 


\begin{table}[!tb]
\small
\centering
\setlength{\tabcolsep}{6mm}
\caption{Classification results on ModelNet40 dataset with different fusion strategies. The view images are rendered under the setup of ~\cite{MVCNN,PVRNet} and the point cloud data are generated following ~\cite{PVRNet}.}
\label{tab:different fusion strategies}
\begin{tabular}{ lcc }
\toprule
Method & OA &mAcc\\
\midrule
Point Cloud Model & 92.2 &  90.2 \\
Multi-View Model & 89.9 & 87.6 \\
\midrule
Late Fusion &  92.6 & 90.8  \\
Our DeepConcat & 93.6 &  91.5\\
View-Point Fusion &  94.0 & 90.9  \\
Point-View Fusion  &  93.8 & 91.4  \\
LATFormer & \textbf{94.4} &  \textbf{92.2}\\
\bottomrule 
\end{tabular}
\end{table}

\begin{table}[t]
\small
\centering
\caption{Classification results under the random point rotation setup. The view images are rendered under the setup of ~\cite{kanezaki2018rotationnet} and the point cloud data are generated following ~\cite{PVRNet}. All the results are reproduced in our environment.}
\label{tab:random_point_rotation}
\vspace{-2ex}
\begin{tabular}{l *{2}{p{0.7cm}<{\centering}} *{2}{p{1.7cm}<{\centering}}}
\midrule
  {\multirow{2}{*}{Method}}&\multicolumn{2}{c}{Aligned}&\multicolumn{2}{c}{Arbitary}\\
  \cline{2-5}
  &OA & mAcc & OA & mAcc\\
\midrule
PVRNet~\cite{PVRNet}  & 97.6 & 96.2 & 97.3 ($\downarrow$ 0.3) & 95.7 ($\downarrow$ 0.5)\\
LATFormer  & \textbf{97.9} & \textbf{97.0} & \textbf{97.4} ($\downarrow$ 0.5) & \textbf{96.0} ($\downarrow$ 1.0)\\
\bottomrule
\end{tabular}
\end{table}

\begin{table}[t]
\small
 \centering
 \setlength{\tabcolsep}{4mm}
\caption{Comparison with the standard transformer on the ModelNet40 dataset under the setup of ~\cite{PVRNet}. The view images are rendered under the setup of ~\cite{MVCNN,PVRNet} and the point cloud data are generated following ~\cite{PVRNet}. $P_t$ represents point cloud data. $I^{\diamond}$ represents view images under setup of ~\cite{MVCNN}. $^{*}$ indicates LATFormer using standard transformer layers for point-view fusion.}
\label{tab:Comparison_with_standard_transformer}
\begin{tabular}{ lccc }
\toprule
Methods & Input & OA & mAcc\\
\midrule
PVNet~\cite{PVNet} &$P_t$, $I^{\diamond}$ & 93.2 & - \\
PVRNet~\cite{PVRNet} &$P_t$, $I^{\diamond}$ & 93.6 & - \\
\midrule
LATFormer-ST$^{*}$ &$P_t$, $I^{\diamond}$ & 94.0 & 91.8 \\
LATFormer(Ours) &$P_t$, $I^{\diamond}$ &\textbf{94.4} & \textbf{92.2} \\
\bottomrule 
\end{tabular}
\end{table}

\noindent\textbf{Robustness to random point rotation.}
In this setting, both training and testing point cloud data are randomly rotated. For the view branch, we adopt 20-view settings following~\cite{kanezaki2018rotationnet, View-GCN} and utilize VGG-11 as the multi-view backbone. For the point branch, we adopt  DGCNN to learn point-wise features. Table~\ref{tab:random_point_rotation} compares the results of PVRNet~\cite{PVRNet} and the proposed LATFormer. From this table, we draw the following conclusions: (1) the proposed LATFormer performs slightly inferior to PVRNet~\cite{PVRNet}. (2) Both methods are robust to point cloud rotation which may benefit from geometric information from different camera views provided by multi-view data.

\noindent\textbf{Comparison with the standard transformer.}
In the standard transformer~\cite{vaswani2017attention, Vit}, the attention weights are normalized by the softmax function while obtained by the thresholded sigmoid function in our method. Table~\ref{tab:Comparison_with_standard_transformer} compares the results of the two methods on the ModelNet40 dataset. As shown, both are superior to previous point-view based methods, which demonstrates the effectiveness of region-to-region relations in the point-view fusion process. Furthermore, our method achieves better results than using standard transformer layers for bi-modal fusion. We assume it is because overnumbered non-corresponding region pairs can easily contaminate the fusion process implemented by the standard transformer while our LATFomer can filter out the low similarity pairs and retain salient local spatial co-occurrence pairs for a sparsified but effective aggregation.


\noindent\textbf{Backbone Networks.} We examine the performance of our method with different point-view backbone networks, which are the combinations of AlexNet~\cite{AlexNet}, VGG-11~\cite{VGG}, DGCNN~\cite{DGCNN} and CurveNet~\cite{curvenet}. As shown in Table~\ref{tab:modelnet_rotaionnet_config}, our fusion strategy can consistently obtain strong results regardless of the backbone networks.
When we choose VGG-11 as view backbone and DGCNN as point backbone, the proposed LATFormer can achieve the best result. 
\begin{table}
\centering
\small
\setlength{\tabcolsep}{0.5mm}
\caption{Comparison with RotationNet~\cite{kanezaki2018rotationnet} and View-GCN~\cite{View-GCN} on ModelNet40 and ModelNet10 datasets. $P_t$ represents point cloud data. $I^{\dagger}$ represents view images under setup of ~\cite{kanezaki2018rotationnet}. OA is reported.}
\label{tab:modelnet_rotaionnet_config}
\begin{tabular}{ cccc }
\toprule
  {\multirow{1}{*}{Method}}&  {\multirow{1}{*}{Input}}&{\multirow{1}{*}{Backbone}}&\multicolumn{1}{c}{ModelNet40}\\
\midrule
RotationNet~\cite{kanezaki2018rotationnet} &$I^{\dagger}$ & AlexNet & 96.4\\
RotationNet~\cite{kanezaki2018rotationnet} &$I^{\dagger}$ &VGG-M & 97.4\\
RotationNet~\cite{kanezaki2018rotationnet} &$I^{\dagger}$ &ResNet-50 & 96.9\\
View-GCN~\cite{View-GCN} &$I^{\dagger}$ & AlexNet & 97.2\\
View-GCN~\cite{View-GCN} &$I^{\dagger}$ & ResNet-18 & 97.6\\
View-GCN~\cite{View-GCN} &$I^{\dagger}$ & ResNet-50 & 97.3\\
\midrule
LATFormer &$P_t$, $I^{\dagger}$ & AlexNet, DGCNN & 97.7\\
LATFormer &$P_t$, $I^{\dagger}$ & VGG-11, DGCNN &\textbf{97.9}\\
LATFormer &$P_t$, $I^{\dagger}$ & AlexNet, CurveNet & 97.4\\
LATFormer &$P_t$, $I^{\dagger}$ & VGG-11, CurveNet & 97.6\\
\bottomrule
\end{tabular}
\end{table}

\noindent\textbf{Influence of different scale combinations.} Table~\ref{tab:comp_dif} studies the effects of different scale combinations. \textcolor{black}{Recall that in our framework, there are three progressive representation scales (emph{i.e.}, $L$=3) for point clouds and multi-view images, respectively. In the Table, ($V_{m}$, $P_{k}$) means fusing $\mathbf{F}_{v,m}$ of view images with $\mathbf{F}_{p', n}$ of point clouds.}
We can observe that with a single scale, the model can already obtain strong performance, reaching 93.8\% and 91.2\% in OA and mAcc, respectively. With more scale features being incorporated, better performance can be obtained. Specifically, with all three scales, we obtain the best results, beating the model using the single scale features by 0.6\% and 1.0\% in OA and mAcc, respectively. 
\textcolor{black}{Moreover, It is also demonstrated that fusing representations at the same corresponding scale can derive slightly better performance (\emph{columns} 3-6) than all the other combinations.} 
These results indicate that more informative features are produced by fusing multi-scale corresponding features since richer relations are exploited. 
\begin{table}[!tb]
\small
\centering
\setlength{\tabcolsep}{7mm}
\caption{Classification results on ModelNet40 dataset under different scale combinations.}
\label{tab:comp_dif}
\begin{tabular}{ lcc }
\toprule
Combinations & OA &mAcc\\
\midrule
      
 \{($V_1$, $P_1$)\} & 93.8 & 91.2\\ 
 \{($V_1$, $P_1$), ($V_2$, $P_2$)\} & 94.0 & 91.6 \\
 \{($V_1$, $P_1$), ($V_2$,$P_2$), ($V_3$,$P_3$)\}  & \textbf{94.4} &  \textbf{92.2}\\
\{($V_1$, $P_2$), ($V_2$, $P_3$), ($V_3$, $P_1$)\} & 94.0 & 91.3 \\ 
\{($V_1$, $P_1$), ($V_2$, $P_3$), ($V_3$, $P_2$)\} & 94.1 & 91.4 \\ 
\{($V_1$, $P_3$), ($V_2$, $P_2$), ($V_3$, $P_1$)\} & 94.0 & 92.0 \\ 
\bottomrule 
\end{tabular}
\end{table}

\noindent\textbf{Influence of different layer features utilized in the view branch.}
Table~\ref{tab:pos_of_map} compares the results of using feature maps from different layers of the AlexNet for extracting multi-scale view features. As shown, adopting the feature maps outputted by the 12$-th$ layer can achieve the best performance. We hypothesize it is because the 12$-th$ layer has a larger receptive field, which provides richer regional context information and facilitates the feature aggregation of co-occurrent local regions.

\noindent\textbf{Influence of $k$.} Table~\ref{tab:nn} shows the impacts of nearest neighbor number $k$ within the SGP layers. As shown, setting the nearest neighbor numbers for each point to 20 gives us the best performance, reaching 94.37\% in OA and 92.15\% in mAcc for classification and 93.05\% for retrieval. It is interesting that further increasing $k$ does not lead to better performance. We assume that too many neighbors may contaminate the local structural information of each point, while fewer neighbors are not sufficient to describe the local geometric information of each point. 

\begin{table}[!tb]
\small
\centering
\setlength{\tabcolsep}{4.5mm}
\caption{Influence of adopting different layer feature maps from AlexNet. The view images are rendered under the setup of ~\cite{MVCNN,PVRNet} and the point cloud data are generated following ~\cite{PVRNet}.}
\label{tab:pos_of_map}
\begin{tabular}{ lccc }
\toprule
Layer number & OA & mAcc & mAP\\
\midrule
 7 & 93.48 & 90.94 & 91.92\\
 9 & 93.71 & 91.32 & 92.15\\
 12 & \textbf{94.37} & \textbf{92.15} & \textbf{93.05}\\
 \bottomrule 
\end{tabular}
\end{table}

\begin{table}[t]
\small
 \centering
 \setlength{\tabcolsep}{6mm}
\caption{Retrieval and classification results on the ModelNet40 dataset when nearest neighbor number $k$ is varying. The view images are rendered under the setup of ~\cite{MVCNN,PVRNet} and the point cloud data are generated following ~\cite{PVRNet}. }
\label{tab:nn}
\begin{tabular}{ lccc }
\toprule
        $k$ & OA & mAcc & mAP\\
        \midrule
        5 &  93.52 & 90.86 & 92.30\\
        10 & 93.64 & 91.75 & 92.58\\
        15 & 93.96 & 91.72 & 92.69\\
        20 & \textbf{94.37} & \textbf{92.15} & \textbf{93.05}\\
        25 & 94.00 & 91.49 & 92.72\\
        30 & 94.00 & 91.81 & 92.81\\
\bottomrule 
\end{tabular}
\end{table}

\begin{table}[t]
\small
\centering
\setlength{\tabcolsep}{6mm}
\caption{Influence of the head numbers in the LAF modules. The view images are rendered under the setup of ~\cite{MVCNN, PVRNet} and the point cloud data are generated following ~\cite{PVRNet}.}
\label{tab:head}
\begin{tabular}{ lccc }
\toprule 
$H$ & OA & mAcc & mAP\\
\midrule
 1 & 93.64 & 90.76 & 92.53\\
 2 & 93.68 & 91.12 & 92.69\\
 4 & 93.88 & 91.50 & 92.42\\
 8 & \textbf{94.37} & \textbf{92.15} & \textbf{93.05}\\
 16 & 93.88 & 91.34 & 92.92\\
 \bottomrule 
\end{tabular}
\end{table}

\begin{table}[!tb]
\small
\centering
\setlength{\tabcolsep}{6mm}
\caption{Retrieval and classification results on ModelNet40 dataset when the value of threshold $\beta$ is varying.}
\label{tab:th}
\begin{tabular}{ lccc }
\toprule
        $\beta$ & OA & mAcc & mAP\\
        \midrule
        0 & 93.80 & 92.00 & 92.85\\
        0.1 & 94.00 & 92.13 & 93.03 \\
        0.3 & \textbf{94.37} & \textbf{92.15} & \textbf{93.05}\\
        0.5 & 93.84 & 91.59 & 92.91\\
        0.7 & 93.72 & 91.62 & 92.64\\
        0.9 & 93.60 & 91.52 & 92.60\\
        \bottomrule 
        \end{tabular}
\end{table}

\noindent\textbf{Influence of $H$.}
Table~\ref{tab:head} presents the performance of our LATFormer with different number of heads in the LAF modules. As shown, with the number of heads increasing, we observe steady improvements in terms of all the evaluation metrics. We believe that with more heads the module can learn richer associations between the two modalities. The best results are obtained when we set $H$ to 8, reaching 94.37\% in mAcc and 92.93\% in mAP. However, when $H$ goes beyond 8, the performance gets saturated. 

\noindent\textbf{Influence of $\beta$.}~The threshold $\beta$ is used to select salient cross-modal co-occurrance for the fusion process. To investigate its effects, we set it to 0, 0.1, 0.3, 0.5, 0.7, 0.9, respectively. Particularly, $\beta = 0$ means not applying a threshold to the scores. As shown in Table~\ref{tab:th}, when we increase $\beta$ from 0 to 0.3, better performance is obtained. However, further increasing it leads to the performance drop.
We assume that if the threshold is too small, then the fused features contain too much redundant information, hurting the fusion efficacy. However, too large value may filter out much valuable information, thus reducing the capacity of the fused features. An appropriate value for the threshold (0.1$\sim$0.3 in our case) benefits the fusion process.

\begin{figure}[!tb] 
\centering 
\includegraphics[width=0.4\textwidth]{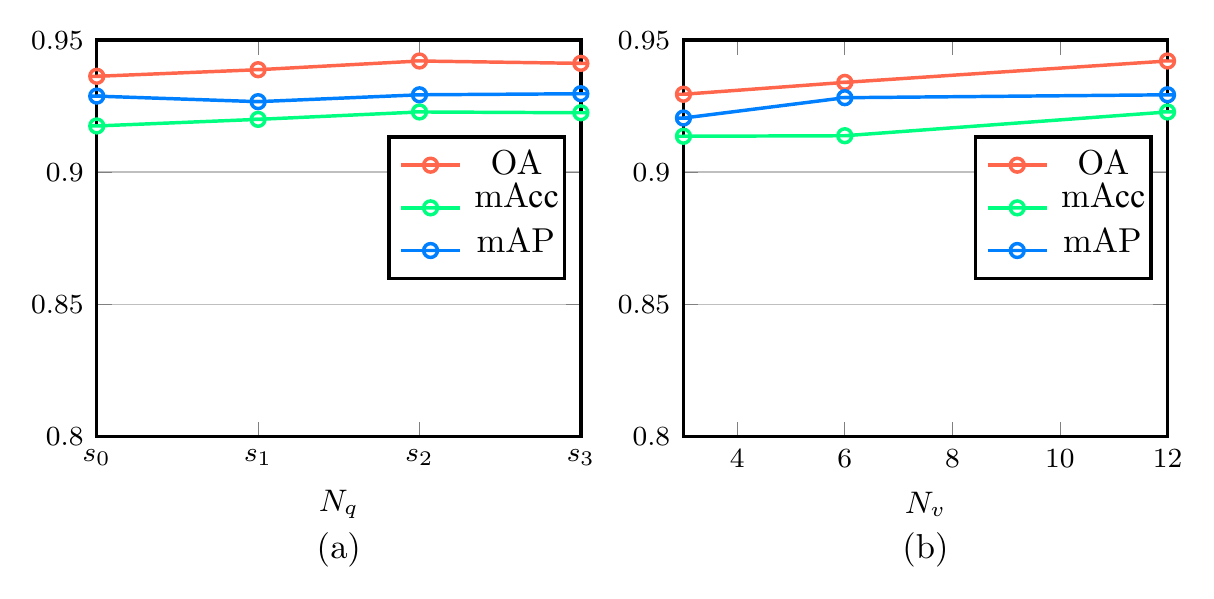} 
\caption{Influence of (a) the number of sampled points and (b) the number of views on the model performance.} 
\label{Fig.view_scale_results} 
\end{figure}

\noindent\textbf{Influence of $N_q$.}~To study the impacts of the sampled point numbers $N_q$ by FPS in the three SGP layers, we first conduct experiments by setting $N_q$ in the three SGP layers to be (128, 64, 32) (denoted as $s_0$) as the base sampled point numbers. 
Then we double the sampled points in all the SGP layers progressively three times (denoted as $s_1$, $s_2$, $s_3$, respectively) and re-run the experiments. As shown in Fig.~\ref{Fig.view_scale_results} (a), the best classification results are achieved when the sampled point numbers are (256, 128, 64). However, further increasing them leads to inferior results. For retrieval, the best performance is achieved when we set the sampled point numbers to (128, 64, 32). 
Generally, our model is robust to the sampled point numbers in the SGP layers to some extent. 

\noindent\textbf{Influence of $N_v$.}~Fig.~\ref{Fig.view_scale_results} (b) studies the influence of the number of views $N_v$ for each 3D object. We vary $N_v$ to be 3, 6, 12 and then conduct experiments to see its impacts. As shown, more views lead to better performance. Specifically, for the classification task, more steady improvements in mAcc and OA are observed when the number of views increases. However, for retrieval, the growth rate of mAP gets smaller after the number of views is beyond 3. It indicates that the classification task benefits more from more view images. 

\begin{table}
\small
\setlength{\tabcolsep}{3mm}
\centering
\caption{Architecture comparison with PVRNet on ModelNet40.}
\label{tab:complexity}
\begin{tabular}{ lcccc }
\toprule 
Methods & Parameters & MACs & OA & mAP\\
\midrule
PVRNet~\cite{PVRNet}  & 70M & 171G & 93.6 & 90.5\\
LATFormer & \textbf{25M} & \textbf{163G} & \textbf{94.4} & \textbf{93.1}\\
\bottomrule 
\end{tabular}
\end{table}

\noindent\textbf{Complexity Analysis.} Table~\ref{tab:complexity} compares the complexity of our LATFormer with the state-of-the-art method PVRNet~\cite{PVRNet}. As shown, the proposed LATFormer has 2.8$\times$ fewer parameters while still outperforms PVRNet~\cite{PVRNet} by a clear margin. Moreover, our LATFormer also reduces the MACs by 8G when the mini-batch size is 16. These results demonstrate the high efficiency of our LATFormer.

\begin{figure}[!tb] 
\centering 
\includegraphics[width=0.38\textwidth]{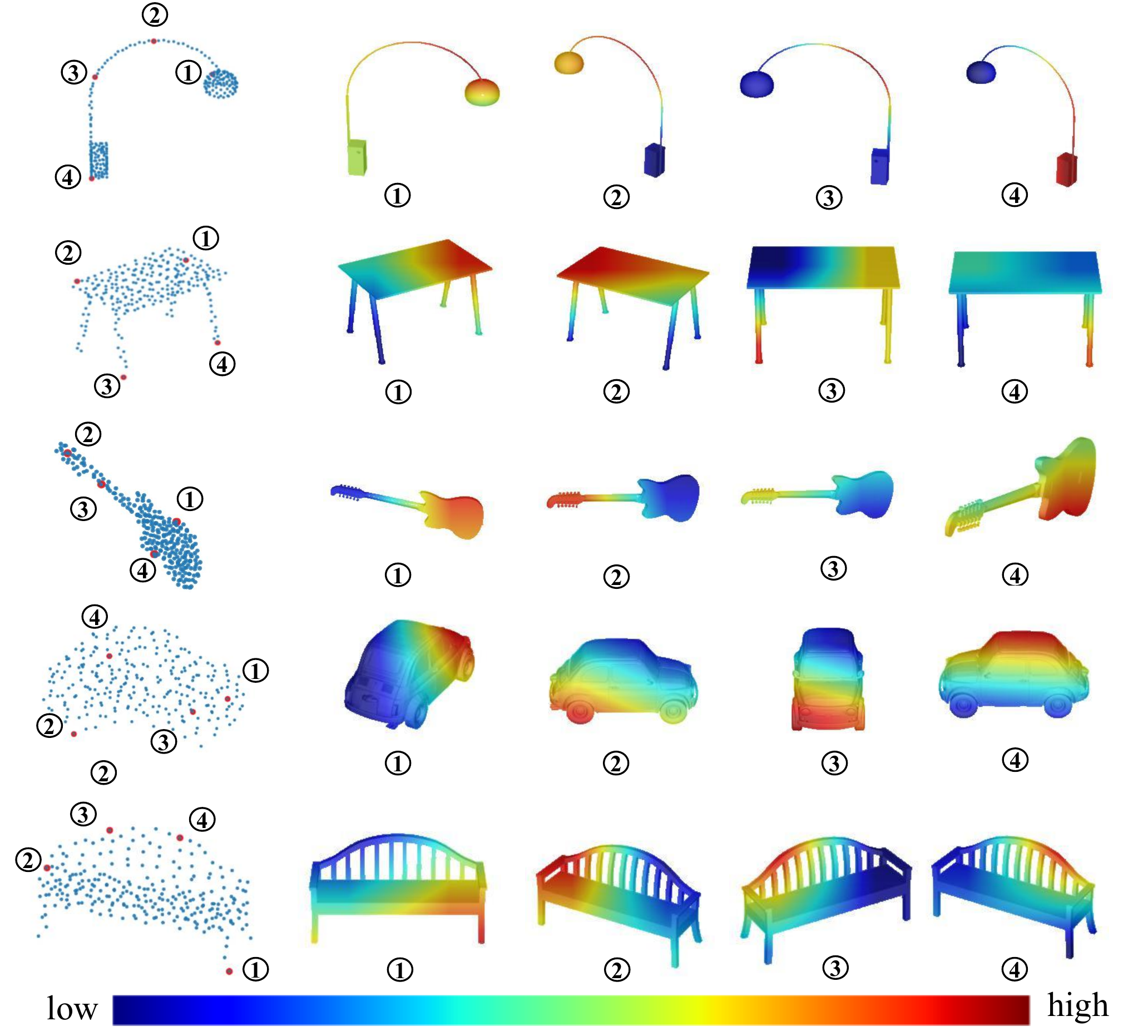} 
\caption{Visualization of the co-occurrance scores learned by the proposed model.} 
\label{Fig.vis_results} 
\end{figure}

\noindent\textbf{Visualization.} We use GradCAM~\cite{selvaraju2017grad} to visualize how much a local region (described by local descriptors at each pixel position of the feature map) in the view images contributes to the related co-occurrence score with a given point, which is shown in Fig.~\ref{Fig.vis_results}. For simplicity, four points are first selected from the point cloud, and then their local co-occurrance scores over the four view images are visualized. It can be observed that LATFormer is highly effective at associating the point features with semantically similar regions in the view images.   


\section{Conclusion and Future Work}\label{sec:conclusion}
In this paper, we have investigated the effects of fusing point cloud and multi-view data based on their local features. A novel module named Locality-Aware Fusion (LAF) is proposed to fuse the two modalities by exploiting their salient local spatial co-occurrance, which are derived by applying the threshold to their co-occurrance scores. With this module, we construct a framework named Locality-Aware Point-view Fusion Net (LATFormer) which works in a bidirectional and hierarchical manner in order to fully leverage their local spatial relations. We have shown that the framework is able to obtain significant improvements over existing methods. In the future, we are interested in applying the LAF module to fuse more modalities of 3D objects such as voxels and meshes to explore its generality. 



\ifCLASSOPTIONcaptionsoff
  \newpage
\fi

{\small
\bibliographystyle{IEEEtran}
\bibliography{references}
}

\vfill

\end{document}